# Beyond Self-Attention: Sub-Quadratic Vision Transformers for Fast Image Captioning

Chiradeep Ghosh
Department of Computer Science and Engineering
National Institute of Technology Durgapur
Durgapur, India
cg.24p10169@nitdgp.ac.in

Dakshina Ranjan Kisku
Department of Computer Science and Engineering
National Institute of Technology Durgapur
Durgapur, India
drkisku.cse@nitdgp.ac.in

***Abstract*—Image captioning is a challenging and significant task aiming to generate coherent and semantically meaningful textual descriptions for given images. To accomplish this task, it requires a deep understanding of visual content along with the ability to express that understanding in natural language. Despite remarkable progress with transformer-based architectures, existing approaches often suffer from limitations, such as a lack of rich local feature representations and the high computational cost of quadratic self-attention. The proposed model focuses on improving computational efficiency by restructuring the vision transformer architecture. In designing this approach, the standard self-attention mechanism in Vision Transformers is replaced with a probabilistic transformer approach based on a Gaussian Mixture Model (GMM), a soft-clustering technique. Instead of computing pairwise attention among all image patches, the model groups similar patches into a fixed number of clusters using an Expectation-Maximization (EM) algorithm. This clustering-based mechanism reduces the computational complexity from quadratic $O(n^2)$ to linear $O(nK)$, where $K << n$. The autoregressive GPT-based decoder is used for caption generation. The model is evaluated on the Flickr 30K dataset, demonstrating competitive and significant improvement over existing works.**

## I. Introduction

In recent years, the rapid growth of visual data has increased the demand for intelligent systems capable of effectively understanding and interpreting images [1]. This demand is particularly significant in domains such as medical imaging [2], autonomous driving systems [3], surveillance [1], and assistive technologies for visually impaired individuals [4], where accurate visual interpretation is essential.

Image captioning aims to automatically generate meaningful natural-language descriptions for images by integrating computer vision and natural language processing techniques. Unlike traditional image classification tasks [5], image captioning requires not only recognizing objects but also understanding their relationships, actions, contextual information, and semantic meaning [6]. Therefore, image captioning has emerged as an important interdisciplinary research area bridging visual understanding [7] and language generation [8].

Early image captioning approaches primarily relied on handcrafted visual features and template-based sentence generation methods [9]. Although these methods were computationally efficient, they often produced rigid and less flexible descriptions. The introduction of deep learning significantly transformed the field through the use of Convolutional Neural Networks (CNNs) [10] for visual feature extraction and Recurrent Neural Networks (RNNs) [11], particularly Long Short-Term Memory (LSTM) networks [12], for sequence generation [13]. More recently, attention mechanisms [14] and Transformer-based architectures [15] have further improved semantic understanding and caption quality.

Modern image captioning systems generally follow an encoder-decoder framework in which a visual encoder extracts semantic image features while a language decoder generates captions sequentially. Researchers continue to explore Transformer-based architectures [15], reinforcement learning techniques [16], attention-based models [7], and vision-language pretraining frameworks [17] to improve contextual reasoning, semantic understanding, and computational efficiency.

Despite substantial progress, generating accurate, contextually meaningful captions remains challenging because it requires a comprehensive understanding of objects, spatial relationships, scene context, and natural-language generation. Nevertheless, image captioning remains an active and evolving research area with significant academic and real-world importance, owing to its potential applications in intelligent visual understanding systems.

Transformer architectures have demonstrated remarkable performance in visual understanding and sequence modelling tasks by capturing long-range contextual dependencies through self-attention [15]. However, the computational complexity of conventional self-attention grows quadratically with the number of input patches, making transformer-based models computationally expensive and memory-intensive, particularly for high-resolution images and large-scale vision-language applications [18].

Additionally, standard self-attention mechanisms often perform redundant pairwise computations between all image patches, even when many patches may not contain semantically important information. This computational redundancy limits scalability and increases inference cost in practical image captioning systems.

To address these challenges, the proposed approach introduces a clustering-based feature extraction mechanism grounded in the Gaussian Mixture Model (GMM) framework [41, 65]. Instead of explicitly computing pairwise attention between all image patches, the proposed architecture models contextual relationships via probabilistic clustering and soft semantic grouping.

Key contributions are summarized as follows.

- The proposed architecture reduces the quadratic time complexity of the conventional self-attention mechanism in the transformer model to near-linear while maintaining the ability to generate semantically meaningful and contextually coherent image captions.

- To maintain computational efficiency, the customised GPT-based architecture is employed as the language decoder. This enhances contextual representation capability
- A comprehensive experimental evaluation of the proposed architecture using widely accepted image captioning metrics has been performed and compared with some existing works.

The rest of the paper is organized as follows. Section II discusses some state-of-the-art image captioning models. Section III presents the proposed methodology that includes image encoding and language decoding modules. The next section reports the evaluation of the proposed model on the benchmark dataset. Concluding remarks and future directions have been presented in the last section.

## II. Literature Survey

Image captioning is a multimodal artificial intelligence task that generates semantically meaningful natural language descriptions of visual content. Over the years, image captioning approaches have evolved from rule-based methods to deep learning and transformer-based architectures.

One of the pioneering works in image captioning was the Show and Tell model proposed by Vinyals et al. [13]. The model employed a Convolutional Neural Network (CNN) to extract visual features from images and a Long Short-Term Memory (LSTM) network to generate captions sequentially. This work achieved remarkable performance on benchmark datasets and established a strong baseline for subsequent research, it relied solely on global image representations and lacked the ability to focus on specific image regions. Moreover, the recurrent nature of LSTM networks limited their capability to capture long-range dependencies effectively.

To overcome the limitations of global feature representations, Xu et al. [14] introduced the Show, Attend and Tell framework, which incorporated an attention mechanism into the image captioning process. The attention module enabled the model to dynamically focus on different regions of an image as it generates each word of the caption. This approach improved caption quality and enhanced interpretability by visualizing the attended image regions. However, the model still relied on recurrent neural networks and could not efficiently model global relationships among all image regions.

Subsequently, Anderson et al. [7] proposed the Bottom-Up and Top-Down Attention model, which became a milestone in image captioning research. The framework utilized object-level features extracted by Faster R-CNN and applied top-down attention during caption generation. By leveraging object proposals instead of grid-based visual features, the model achieved superior performance on the MS-COCO benchmark. However, the framework relied heavily on external object detectors, which increased computational complexity and introduced error propagation from the detection.

The emergence of Transformer architectures transformed the landscape of image captioning. Inspired by the success of the Transformer in natural language processing, Parmar et al. [19] introduced the Image Transformer, which utilized self-attention mechanisms to capture long-range dependencies in image representations. The model demonstrated the ability of attention-based architectures to model global contextual relationships more effectively. Despite these advantages, the self-attention mechanism exhibited quadratic computational complexity with respect to sequence length, leading to substantial memory and computational requirements.

A major breakthrough in visual representation learning was achieved with the introduction of the Vision Transformer (ViT) by Dosovitskiy et al. [20]. ViT treated images as sequences of image patches and processed them using standard Transformer encoders. The model demonstrated that pure Transformer architectures could outperform traditional CNNs when trained on sufficiently large datasets. ViT effectively captured global contextual information and provided highly expressive visual representations. However, its reliance on full self-attention resulted in quadratic complexity and substantial computational overhead, particularly for high-resolution images.

To address the computational limitations of standard self-attention, LINFORMER [21] introduced by Wang et al. reduced attention complexity through low-rank projections of the key and value matrices. This approach significantly decreased memory consumption and computational cost while maintaining competitive performance. However, the low-rank approximation could result in information loss and reduced representation quality in certain scenarios. Similarly, Reformer, proposed by Kitaev et al. [22], employed locality-sensitive hashing (LSH) to approximate self-attention and improve scalability. Although REFORMER reduced memory requirements substantially, the approximate attention mechanism sometimes failed to capture important dependencies.

Another notable efficient Transformer variant is PERFORMER, proposed by Choromanski et al. [23]. PERFORMER introduced the FAVOR+ mechanism, approximating softmax attention via kernel-based random feature mappings. This technique achieved linear attention complexity while preserving much of the effectiveness of standard attention. However, approximation errors and sensitivity to hyperparameter selection could negatively affect performance in certain applications.

Recent advances in image captioning have increasingly leveraged large language models for text generation. Models such as ClipCap [24], Frozen [25], and BLIP [26] integrate powerful language models with visual encoders to generate highly fluent and contextually rich captions. However, these systems often require extensive pretraining data, large computational resources, and high memory capacity, making them expensive to train and deploy.

Motivated by these limitations, the proposed work presents an improved GMM-based Transformer visual encoder that replaces conventional self-attention with a Gaussian Mixture Model (GMM)-based soft-clustering mechanism [27]. By grouping visually similar tokens into clusters and performing an attention equivalent operation at the cluster level, the proposed encoder reduces computational complexity from quadratic to semi-quadratic while preserving important contextual relationships. The generated visual representations are subsequently provided to an autoregressive GPT-based Transformer decoder for language generation. By combining efficient visual encoding with powerful autoregressive language modeling, the proposed framework seeks to achieve an improved balance between caption quality, computational efficiency, and scalability in image captioning applications.

## III. Proposed Methodology

In this work, the overall architecture is inspired by the Vision Transformer (ViT) framework. Core transformer components, including patch embeddings, positional encodings, feed-forward networks, and layer normalization, are retained within the architecture. However, the conventional multi-head self-attention module is replaced with a GMM-based probabilistic clustering block for contextual feature extraction and encoding. For language decoding and image caption generation, an autoregressive GPT-based Transformer is used. The proposed image captioning model is shown in Figure 1.

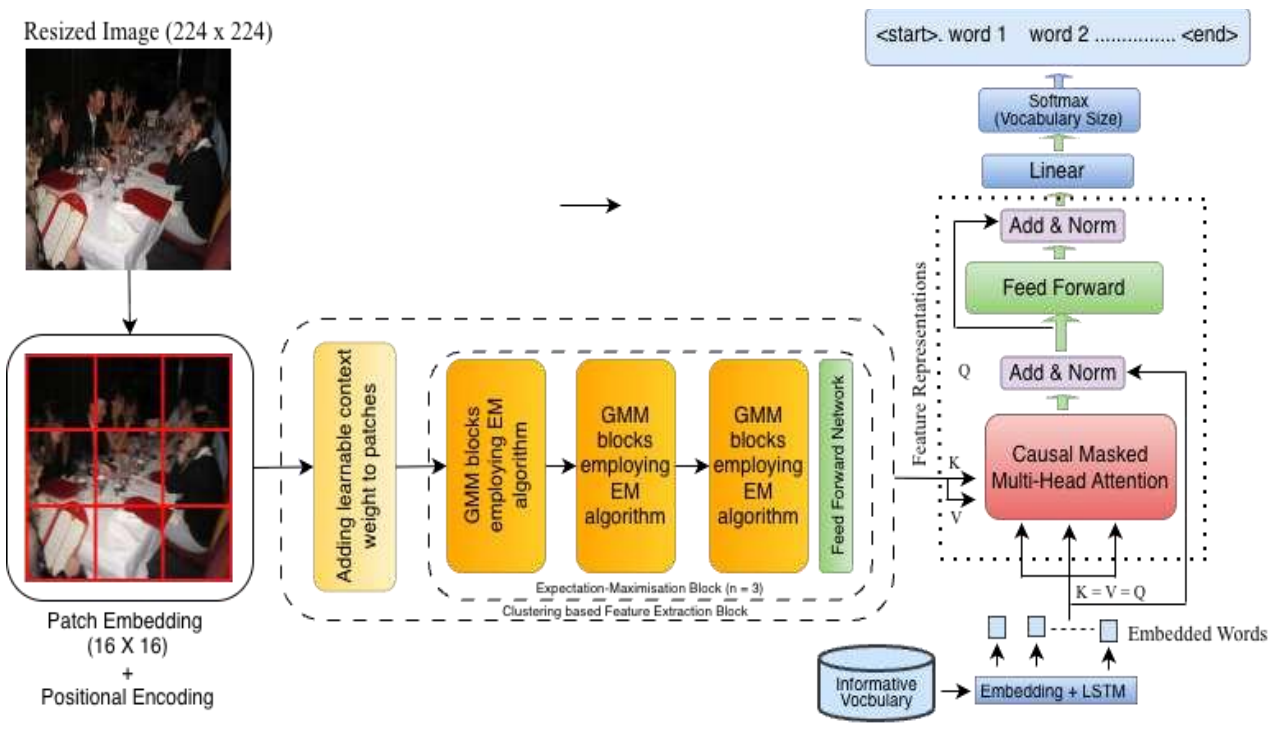

Fig. 1. Proposed Image Captioning Model

### A. Image Encoding

The image encoder is designed to capture contextual image features and dependencies while maintaining computational efficiency.

Let an input image $I$ be divided into $N$ non-overlapping patches, each linearly projected to a $D-$ dimensional embedding. Each patch represents a local region of the image. This process is performed using a patch embedding module, similar to that used in Vision Transformer architectures. To preserve spatial relationships, positional encodings are added to the patch embeddings. This ensures that the model retains information about the original spatial arrangement of the patches. The patch embeddings are represented as:

$$X = [x_1, x_2, \ldots, x_N]^T \in \mathbb{R}^{N \times D}$$

The standard Vision Transformer initially computes query, key, and value vectors using patch embeddings and learnable weights, and then uses these weighted embeddings to compute attention scores within the self-attention paradigm. The query, key and value, and attention score are given by

$$Q = XW_Q, \quad K = XW_K, \quad V = XW_V \tag{1}$$

$$Attention(Q, K, V) = softmax(\frac{QK^T}{\sqrt{D}})V \tag{2}$$

This computation is performed with computational complexity $O(N^2 D)$.

Instead of traditional scaled Dot-Product attention, we model the patch embeddings $X = [x_1, x_2, \ldots, x_N]$ as continuous random variables generated by a mixture of $K$ multivariate Gaussian distributions, where $K \ll N$. The positionally encoded patches are forwarded to a sequence of transformer-like blocks, where the conventional pair-wise self-attention mechanism is replaced with a Gaussian Mixture Model (GMM)-based soft-clustering module followed by cluster-level attention. The feature encoding module with the GMM-Transformer layer is shown in Figure 2.

Before clustering, a Learnable Context Weights sub-layer is applied. This layer generates a context-weight tensor for each patch. It learns a spatial weight matrix, where each patch location is associated with a learnable weight vector. These weights are further modulated by the statistical properties of the patches, such as their mean and variance. This mechanism enables the model to assign varying importance to different spatial regions of the image. For instance, certain regions may contribute more significantly to semantic understanding than others. These weights are optimized end-to-end through backpropagation.

The output of this sub-layer, along with the normalized patch embeddings and the original image representation, is passed to the Context-Weighted GMM Clustering module. This module performs soft clustering of patches based on feature similarity, guided by the learned context weights.

To obtain a probabilistic encoder which is presented as a GMM-Transformer layer, we define the marginal probability density function of a patch $x_i$ is as follows:

$$p(x_i \mid \Theta) = \sum_{k=1}^{K} \pi_k \mathcal{N}(x_i \mid \mu_k, \Sigma_k) \tag{3}$$

Where the parameter set $\Theta = \{\pi_k, \mu_k, \Sigma_k\}_{k=1}^{K}$:

- $\pi_k$: Mixing coefficients where $\sum_{k=1}^{K} \pi_k = 1$ and $\pi_k \geq 0$.
- $\mu_k \in \mathbb{R}^D$: Mean vector of the $k^{th}$ cluster.
- $\Sigma_k \in \mathbb{R}^{D \times D}$: Covariance matrix of the $k^{th}$ cluster.

The clustering process is implemented using the Expectation-Maximization (EM) algorithm. In the Expectation (E-step), the model computes soft assignment probabilities (responsibilities) that indicate the degree to which each patch belongs to each cluster. In the Maximization (M-step), cluster parameters such as means, covariances, and mixture coefficients are updated based on these responsibilities.

Through iterative EM updates, patches that share similar semantic characteristics are grouped together. This results in a refined representation where each patch is influenced by the cluster to which it belongs. Consequently, patches within the same semantic group reinforce each other's feature representations, enhancing the model's ability to capture meaningful relationships within the image.

After clustering, the refined patch embeddings are processed by a feed-forward network comprising two fully connected layers with GELU activations and dropout regularization. Residual connections are applied around both the clustering module and the feed-forward network, and Layer Normalization is applied prior to each sub-layer, following the standard pre-normalization transformer design.

Multiple such blocks (e.g., N = 4) are stacked sequentially. As the representations propagate through these blocks, the patch features are progressively refined. In the initial stages, the clustering module's contribution is limited by a learnable scaling parameter, leading the model to rely more on residual connections. As training progresses, the influence of clustering increases, enabling stronger incorporation of semantic grouping information.

After passing through all transformer blocks, a final Layer Normalization is applied to the refined patch representations. These patch features are then aggregated using a Global Average Pooling operation across all patch positions to obtain a compact global feature vector representing the entire image.

This feature vector encodes spatial structure, contextual dependencies, and clustering-based relationships within the image. It can subsequently be used for generating descriptions for the images.

Let's consider, for each encoder layer, the EM algorithm runs for a fixed number of internal iterations $T$ to estimate the latent cluster assignments.

*a) E – Step (Responsibility Estimation):*

We now compute the posterior probability (responsibility) that the cluster $k$ generated patch $x_i$:

$$\gamma_{ik} = p(z_i = k \mid x_i, \Theta) = \frac{\pi_k \mathcal{N}(x_i \mid \mu_k, \Sigma_k)}{\sum_{j=1}^{K} \pi_j \mathcal{N}(x_i \mid \mu_j, \Sigma_j)} \quad (4)$$

This matrix $\Gamma \in \mathbb{R}^{N \times K}$ acts as a soft-clustering assignment matrix and replaces the traditional $N \times N$ attention matrix.

*b) M – Step (Parameter Update):*

We update the mixture parameters using the calculated responsibilities:

$$N_k = \sum_{i=1}^{N} \gamma_{ik} \quad (5)$$

$$\pi_k^{new} = N_k / N \quad (6)$$

$$\mu_k^{new} = \frac{1}{N_k} \sum_{i=1}^{N} \gamma_{ik} x_i \quad (7)$$

$$\Sigma_k^{new} = \frac{1}{N_k} \sum_{i=1}^{N} \gamma_{ik} (x_i - \mu_k^{new})(x_i - \mu_k^{new})^T \quad (8)$$

The log-likelihood of the data under the model is evaluated as:

$$\log L = \sum_{n=1}^{N} log \left( \sum_{k=1}^{K} \pi_k \mathcal{N}(x_n \mid \mu_k, \Sigma_k) \right) \quad (9)$$

The Expectation–Maximization procedure is repeated until convergence is achieved, typically when the improvement in the log-likelihood falls below a predefined threshold, or the maximum number of iterations is reached.

*c) Encoding Function*

The final encoded representation of the image is the optimized matrix of cluster centroids scaled by their structural presence:

$$Z = [\mu_1, \mu_2, \dots, \mu_K]^T \in \mathbb{R}^{K \times D}$$

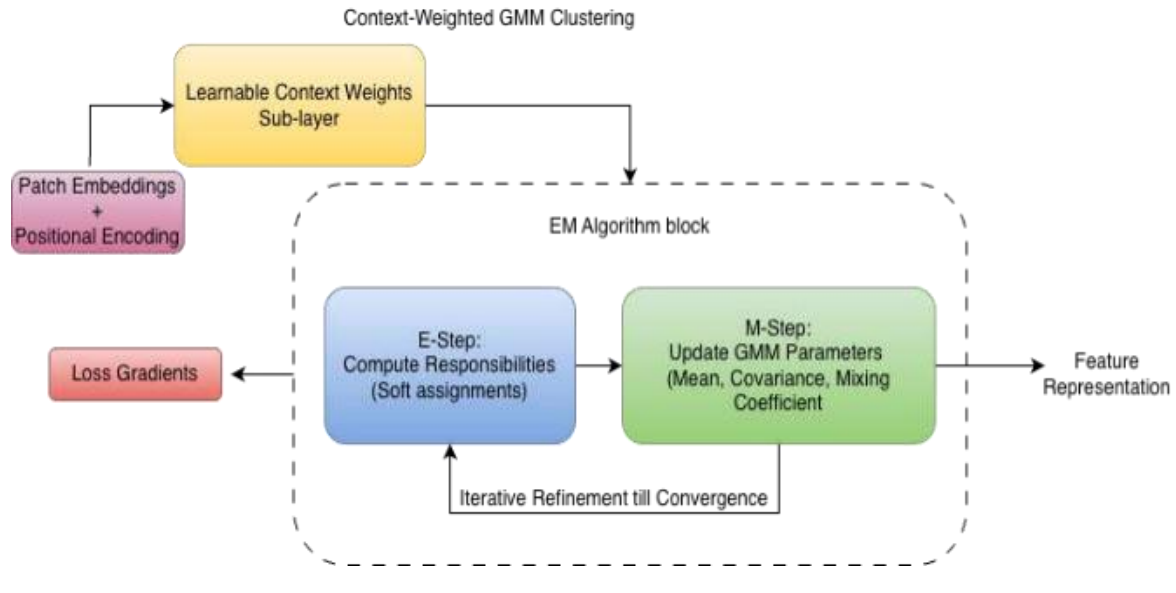

Fig. 2. Feature Representation with GMM-Transformer Layer

*B. Language Decoding*

The proposed decoder architecture is based on a unified GPT-style autoregressive transformer framework designed specifically for multimodal image caption generation. Unlike conventional encoder–decoder transformer architectures, which implement masked self-attention and cross-attention as separate computational modules, the proposed decoder integrates both mechanisms into a single multi-head attention block to improve computational efficiency and multimodal interaction.

After the image encoder extracts rich local and global visual representations, the image features are concatenated with embedded textual tokens and then processed by the decoder. The textual inputs are first transformed into dense vector embeddings with positional encodings to preserve sequential order. During caption generation, a specialized causal masking strategy is employed, ensuring that each textual token can attend only to previously generated tokens while simultaneously attending to all visual tokens extracted from the image encoder.

In this unified attention formulation, textual tokens serve as query vectors, while the combined visual and textual representations serve as key and value vectors. This enables the decoder to jointly learn linguistic dependencies and vision-language correlations within a single attention computation, thereby reducing computational redundancy present in traditional architectures. The attention outputs are further refined via residual connections, layer normalization, and feed-forward neural networks before being passed through a Soft-max layer to generate a probability distribution over the vocabulary. The most probable token is then selected iteratively and fed back into the decoder in an autoregressive manner until the end-of-sequence token is generated. By consolidating masked self-attention and cross-attention into a unified framework, the proposed GPT-based decoder achieves efficient multimodal feature fusion while generating semantically coherent, contextually meaningful, and grammatically accurate image captions.

*a) Decoder Input Sequence Fusion*

Let $Z = [\mu_1, \mu_2, \dots, \mu_K]^T \in \mathbb{R}^{K \times D}$ be the sequence of $K$ cluster centroids generated by the GMM-ViT image encoder and $W = (w_1, w_2, \dots, w_t)$ be the sequence of $t$ generated text tokens at the current decoding step. These tokens are mapped to a dense space and injected with positional encodings $E_{pos} \in \mathbb{R}^{t \times D}$:

$$H_{text}^{(0)} = [h_1, h_2, \dots, h_t]^T \in \mathbb{R}^{t \times D} \quad (10)$$

$$h_i = Embeddings(w_i) + E_{pos,i} \quad (11)$$

The visual tokens and textual embeddings are then concatenated along the sequence dimension to form a unified input matrix $U \in \mathbb{R}^{(K+t) \times D}$:

$$U = [Z; H_{text}^{(0)}] = [\mu_1, \mu_2, \dots, \mu_K, h_1, h_2, \dots, h_t]^T \quad (12)$$

*b) Unified Multimodal Multi-Head Attention*

The unified matrix $U$ and the text token representations $H_{text}$ are projected into Query $(Q)$, Key $(K)$, and Value $(V)$ spaces using learnable weights $W_Q$, $W_K$, $W_V \in \mathbb{R}^{D \times D}$. To enforce the architecture where text tokens act as queries while the combined sequence acts as keys and values:

$$Q = H_{text}^{(0)} W_Q \in \mathbb{R}^{t \times D} \quad (13)$$

$$K_v = UW_K \in \mathbb{R}^{(K+t)\times D} \quad (14)$$

$$V = UW_V \in \mathbb{R}^{(K+t)\times D} \quad (15)$$

Now, to allow full visibility over image tokens but strict autoregressive visibility over text tokens, a causal mask matrix $M \in \mathbb{R}^{t\times(K+t)}$ is constructed as follows:

$$M_{i,j} = \begin{cases} 0, \ if\ j \leq K\ (Visual\ Tokens) \\ 0, \ if\ K < j \leq K+i\ (Past/Current\ Text\ Tokens) \\ -\infty, \ if\ j > K+i\ (Future\ Text\ Tokens) \end{cases} \quad (16)$$

Visual representation of the mask matrix $M$ is given by

$$M = \begin{pmatrix} 0 & \dots & 0 & 0 & -\infty & \cdots & -\infty \\ 0 & \dots & 0 & 0 & 0 & \cdots & -\infty \\ \vdots & \ddots & \vdots & \vdots & \vdots & \ddots & \vdots \\ 0 & \dots & 0 & 0 & 0 & \cdots & 0 \end{pmatrix} \quad (17)$$

On the use of the mask matrix, the integrated self- and cross-attention output for a single head is computed as

$$Attention(Q, K_v, V) = Softmax\left(\frac{QK_v^T}{\sqrt{D}} + M\right)V \in \mathbb{R}^{t\times D} \quad (18)$$

and for $H$ parallel attention heads, the outputs are concatenated and linearly projected via $W_O \in \mathbb{R}^{D\times D}$:

$$MultiHead(Q, K_v, V) = Concat(head_1, \dots, head_H)W_O \quad (19)$$

*c) Layer Processing and Feature Refinement*

In the next step, the output of the unified multi-head attention block passes through residual connections, layer normalization (LN), and a Position-Wise Feed-Forward Network (FFN):

$$H_{attn} = LN(H_{text}^{(0)} + MultiHead(Q, K_v, V)) \quad (20)$$

$$H_{text}^{(1)} = LN(H_{attn} + FFN(H_{attn})) \quad (21)$$

Where the FFN is defined using two linear transformations and an activation function as follows:

$$FFN(x) = \max(0, xW_1 + b_1)\,W_2 + b_2 \quad (22)$$

*d) Autoregressive Vocabulary Projection*

The final hidden state of the latest text token at index $t$, denoted as $h_t^{(L)}$ from the final decoder layer $L$, is projected onto the vocabulary space $\mathcal{V}$:

$$z_t = h_t^{(L)}W_{vocab} + b_{vocab} \in \mathbb{R}^{|\mathcal{V}|} \quad (23)$$

The probability distribution over the entire vocabulary space for the next token $w_{t+1}$ is computed via the Softmax function:

$$P(w_{t+1}\,|w_{\leq t}, Z) = Softmax(z_t)_i = \frac{e^{z_{t,j}}}{\sum_{j=1}^{|\mathcal{V}|} e^{z_{t,j}}} \quad (24)$$

The token sequence is updated iteratively by selecting the most probable token index until the end-of-sequence token is generated:

$$w_{t+1} = arg\max_i P(w_{t+1} = i\,|w_{\leq t}, Z) \quad (25)$$

## IV. Evaluation

The evaluation of the proposed image captioning architecture is performed on the Flickr30k dataset [28]. Prior to training, preprocessing steps are employed to improve semantic consistency and training efficiency.

The proposed model is trained within an encoder-decoder framework using the Google Colab L4 GPU environment. The model is optimized using the AdamW optimizer and the Sparse Categorical Cross-Entropy (SCCE) loss function, where caption generation is formulated as a multi-class classification problem over the vocabulary space.

To analyze model convergence and learning behaviour, training and validation loss curves are determined across epochs for the proposed model. Similarly, training and validation accuracy plots are analysed to study generalisation performance and learning stability. Confusion matrix analysis is also performed using predicted and ground-truth words to compute True Positives (TP), False Positives (FP), False Negatives (FN), and True Negatives (TN).

In addition, computational efficiency is analyzed using GFLOPs and trainable parameter count. The qualitative performance of the proposed architecture is evaluated using generated captions for several sample images from the Flickr30k dataset. The model generates richer, more contextually detailed captions through its hybrid local-global feature extraction mechanism, with significantly reduced computational complexity. For quantitative evaluation, several standard image captioning metrics [8], [9], [28], [29] are employed, including BLEU, METEOR, ROUGE, CIDEr, and SPICE. The proposed model is also compared with some state-of-the-art image captioning models.

### A. Dataset and Preprocessing

The Flickr30k dataset is specifically designed for image captioning tasks and consists of 31,783 images. Each image is annotated with five distinct captions, resulting in a total of 158,915 textual descriptions. For experimental consistency, the dataset is partitioned using the Karpathy split [29], which splits the dataset into 29,000 images for training, 1,000 for validation, and 1,000 for testing.

Prior to model training, a series of preprocessing steps is applied to both images and their corresponding captions to improve learning efficiency. After data organization, the textual captions undergo a cleaning process to improve their quality and consistency. This includes removing punctuation marks, isolated single-character tokens, and numerical terms that do not contribute meaningful semantic information. Additionally, efforts are made to reduce ambiguity and standardise the textual format. Finally, the cleaned captions are tokenized to convert the text into sequences of tokens, which can be effectively processed by the neural network during training.

### B. Model Parameters

Table 2 enlists the architectural hyperparameters and training configurations for the image-encoding and description-generating model. Input images are processed at 224×224×3 resolution and split into 16×16 patches. The core architecture uses a Transformer with 8 layers, 4 attention heads, a hidden dimension ($d_{model}$) of 768, and an embedding layer of the same size. Image feature map dimensions are set to 1024, and pixel clustering configurations range from 2 to 100 clusters, with 10 iterations for EM and k-means algorithm. For the training pipeline, the model uses a batch size of 32, a buffer size of 1000, a dropout rate of 0.1, and an early-stopping patience of 2 epochs. Optimization is handled with the AdamW optimizer and a custom learning rate scheduler, and performance is evaluated using the Sparse Categorical Cross-Entropy (SCCE) loss function. The setup comprises 7,41,120 patch embedding parameters and 1,536 layer normalisation parameters, totalling 32,793,936 trainable parameters.

TABLE I. HYPERPARAMETERS AND TRAINING CONFIGURATIONS

| Hyperparameter | Dimension/Value |
|---|---|
| Image height x width x channel | 224 x 224 x 3 |
| Patch height and width | 16 |
| Transformer Layers | 8 |
| Context dim | 256 |
| Max no of Pixel Clusters | 100 |
| Min no of Pixel Clusters | 2 |
| EM-iterations | 10 |
| k-means iteration | 10 |
| Number of transformer heads | 4 |
| Feature map dimension | 1024 |
| Embedding layer | 768 |
| Batch size | 32 |
| dmodel | 768 |
| Buffer size | 1000 |
| Early stopping | Patience = 2 |
| Learning rate | Custom-LR Scheduler |
| Dropout | 0.1 |
| Linear Layer | 512 |
| FC Layer 2 | Vocabulary size |
| Optimizer | AdamW |
| Loss Function | SCCE |
| Patch Embed Layer Params | 741,120 |
| Layer Norm Params | 1,536 |
| Total trainable params | 32,793,936 |

### C. *Experimental Results*

Figure 4 presents sample captions generated by the proposed model for selected images from the Flickr30k dataset. These examples illustrate the model's ability to produce coherent, semantically meaningful, and contextually relevant descriptions of visual content and to significantly reduce computational time complexity by eliminating the self-attention mechanism of ViTs with the GMM algorithm. The generated captions demonstrate that the model effectively captures object-level details and the relationships among elements within the image, while maintaining a significantly low computational cost.

This prediction process can be interpreted as a multi-class classification problem over the vocabulary space. To quantify the prediction error, the Categorical Cross-Entropy loss function is used. Using this loss function, the training and validation loss values are computed across epochs. The corresponding loss curve, illustrating the model's convergence behaviour, is shown in Figure 3(a).

The accuracy across different training epochs is illustrated in Figure 3(b), which shows the progression of both training and validation accuracy. This visualisation provides insight into the model's learning behaviour and generalisation performance over time.

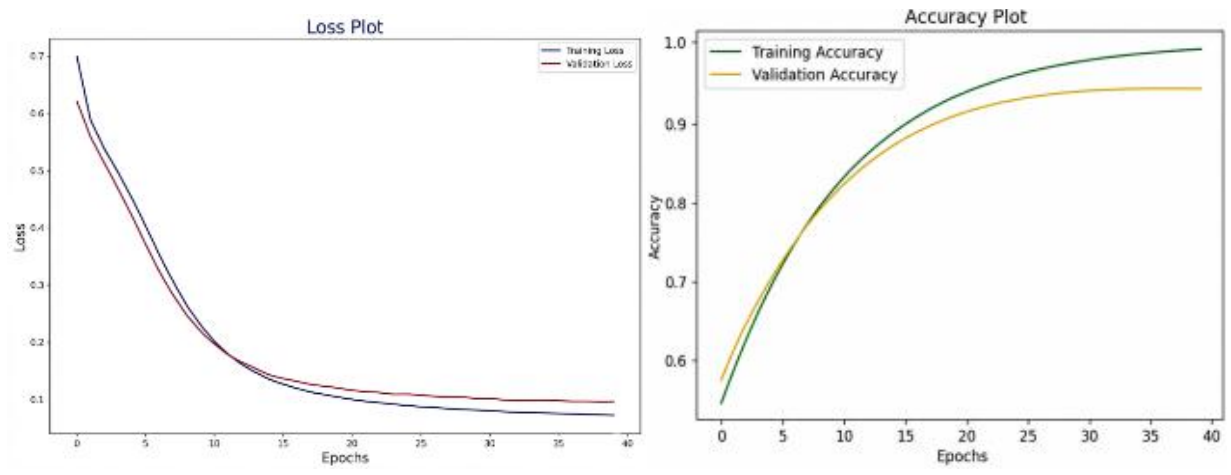

Fig. 3. Loss and Accuracy Graphs

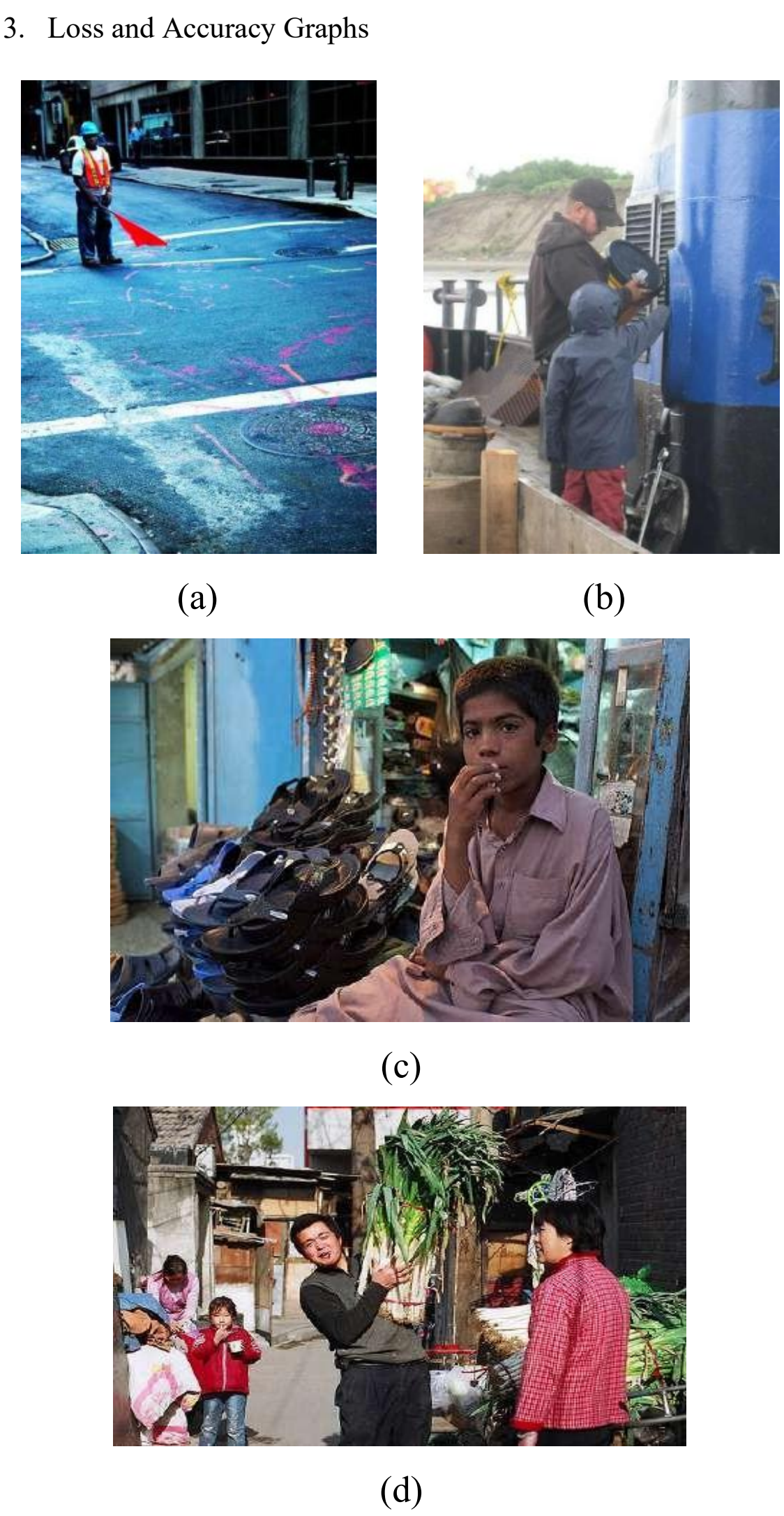

Fig. 4. Qualitative Results: (a) **Real Caption:** A man in a blue hard hat and orange safety vest stands in an intersection while holding a flag. **Predicted Caption:** A man wearing a reflective vest and a hard hat holds a flag in the road. (b) **Real Caption:** A father and son working on a boat. **Predicted Caption:** A man and a child stand next to a blue and black tube. (c) **Real Caption:** A young dark-skinned boy in a large shirt sitting next to a large pile of sandals. **Predicted Caption:** A young boy sits beside a stack of sandals. (d) **Real Caption:** A man in a market holding some very large green onions. **Predicted Caption:** A man in all black carries a vegetable smoking

Table II indicates that the model achieves 45,580 true positives and 17,186,403 true negatives, while producing 3,583 false positives and 4,434 false negatives at a computational cost of 21.17 GFLOPs. These results suggest that the model is highly effective in distinguishing between positive and negative classes, with the large number of correct predictions reflecting strong overall classification performance. The relatively low false positive and false negative counts indicate good precision and recall, meaning the model can identify positive instances accurately while minimizing misclassifications. Furthermore, the computational requirement of 21.17 GFLOPs is moderate for modern deep learning systems. The model demonstrates strong predictive capability.

TABLE II. CONFUSION MATRIX COMPONENTS AND OVERHEAD

| Metric | Value |
| --- | --- |
| True Positives (TP) | 45580 |
| False Positives (FP) | 3583 |
| False Negatives (FN) | 4434 |
| True Negatives (TN) | 17186403 |
| GFLOPs | 21.1749 |

The experimental results of the proposed image captioning model are shown in Table III, and a comparison with state-of-the-art (SOTA) models is presented in the same table. On the Flickr 30k dataset, the proposed model sets new semantic benchmarks. It achieves top scores in BLEU-4 (0.364), METEOR (0.368), and SPICE (0.199). These results prove superior phrase structure and scene graph mapping.

TABLE III. PERFORMANCE OF THE PROPOSED IMAGE CAPTIONING MODEL AND COMPARISON WITH SOTA MODELS

| Model | B-1 | B-2 | B-3 | B-4 | M | R | C | S |
| --- | --- | --- | --- | --- | --- | --- | --- | --- |
| TAKG [30] | 0.683 | – | – | 0.265 | 0.219 | 0.480 | 0.575 | 0.163 |
| GNN-Visual [31] | 0.752 | 0.541 | 0.406 | 0.322 | 0.245 | 0.525 | 0.646 | – |
| Scene Graph [32] | – | – | – | 0.338 | 0.243 | 0.479 | 0.682 | 0.184 |
| AoA [33] | 0.728 | – | – | 0.297 | 0.223 | 0.501 | 0.643 | 0.166 |
| High-Order [34] | 0.732 | – | – | 0.301 | 0.229 | 0.508 | 0.650 | 0.168 |
| Multimodal [35] | 0.758 | 0.587 | 0.445 | 0.333 | 0.245 | – | – | – |
| SCST [36] | 0.734 | – | – | 0.301 | 0.226 | – | 0.693 | 0.168 |
| GAT [37] | 0.744 | 0.567 | 0.418 | 0.308 | 0.234 | - | 0.680 | – |
| TBDATN [38] | 0.765 | 0.589 | 0.445 | 0.295 | 0.258 | 0.602 | 0.702 | – |
| TopDown [39] | 0.737 | – | – | 0.308 | 0.232 | – | 0.706 | 0.176 |
| RFT [40] | 0.716 | – | – | 0.306 | 0.237 | 0.516 | 0.733 | 0.174 |
| DRET [41] | 0.705 | 0.527 | 0.390 | 0.288 | 0.226 | 0.498 | 0.636 | – |
| ADA [42] | 0.794 | 0.596 | 0.437 | 0.324 | 0.283 | 0.608 | 0.946 | 0.188 |
| Proposed Model | 0.733 | 0.438 | 0.267 | 0.364 | 0.368 | 0.568 | 0.723 | 0.199 |

However, it trails the ADA model in baseline precision. ADA leads in BLEU-1 (0.794), ROUGE (0.608), and CIDEr (0.946). The proposed model marks 0.733, 0.568, and 0.723, respectively. Yet it still outpaces baselines such as TopDown and GNN-Visual. Clearly, the architecture favours deep context over simple replication. The proposed model significantly outperforms all state-of-the-art (SOTA) models in advanced linguistic and semantic metrics (BLEU-4, METEOR, and SPICE), while remaining highly competitive in baseline precision and recall metrics (BLEU-1, ROUGE, and CIDEr).

### D. Computation Complexity

In particular, the transformer-based self-attention mechanism computes pairwise interactions among all image patches, resulting in quadratic time complexity $O(n^2)$, where $n$ denotes the number of image patches.

In contrast, the proposed model has achieved substantial reduction in both model size and storage requirements. The reduction in parameter complexity is primarily achieved by replacing the conventional self-attention mechanism with a Gaussian Mixture Model (GMM)-based probabilistic clustering framework. Instead of computing attention relationships among all image patches, the proposed probabilistic transformer groups semantically similar patches into clusters and performs contextual aggregation using probabilistic assignments. This modification reduces the computational complexity from quadratic complexity $O(n^2)$ to approximately near-linear complexity $O(nk)$, where $k \ll n$ and $k$ denotes the number of clusters.

## V. CONCLUSION AND FUTURE WORKS

The proposed model addresses the computational inefficiency associated with the self-attention mechanism in Vision Transformer-based architectures. Specifically, it replaces the conventional self-attention module with a clustering-based approach derived from the Gaussian Mixture Model (GMM) and its Expectation-Maximization algorithm. This modification significantly reduces the computational complexity from quadratic to near-linear, while still maintaining the ability to capture meaningful feature relationships. The model exhibits strong predictive capability, scalable feature encoding and outperforms SOTA models on BLEU-4, METEOR, and SPICE scores. For other performance metrics, the model performs competitively with the existing models.

Future research aims to enhance image captioning systems by incorporating advanced multimodal pretraining, adapting architectures for complex, domain-specific datasets, and utilizing reinforcement learning to improve caption quality. Additionally, focus areas include deploying lightweight, adaptive clustering models for real-time edge applications, supporting multilingual generation, and improving interpretability through attention visualization.